# A Study on Artificial Intelligence IQ and Standard Intelligent Model

Feng Liu[a*], Yong Shi[b,c,d]

[a]School of Computer and Information Technology, Beijing Jiaotong University, Beijing 100044, China

[b]Research Center on Fictitious Economy and Data Science, the Chinese Academy of Sciences, China, Beijing 100190, China

[c]The Key Laboratory of Big Data Mining and Knowledge Management, Chinese Academy of Sciences, China, Beijing 100190, China

[d]Collegae of Information Science and Technology, University of Nebraska at Omaha, Omaha, NE 68182, USA

## Abstract

Currently, potential threats of artificial intelligence (AI) to human have triggered a large controversy in society, behind which, the nature of the issue is whether the artificial intelligence (AI) system can be evaluated quantitatively. This article analyzes and evaluates the challenges that the AI development level is facing, and proposes that the evaluation methods for the human intelligence test and the AI system are not uniform; and the key reason for which is that none of the models can uniformly describe the AI system and the beings like human. Aiming at this problem, a standard intelligent system model is established in this study to describe the AI system and the beings like human uniformly. Based on the model, the article makes an abstract mathematical description, and builds the "standard intelligent machine" mathematical model; expands the Von Neumann architecture and proposes the Liufeng - Shiyong architecture; gives the definition of the artificial intelligence IQ, and establishes the "artificial intelligence scale" and the evaluation method; conduct the test on 50 search engines and three human subjects at different ages across the world, and finally obtains the ranking of the absolute IQ and deviation IQ ranking for artificial intelligence IQ 2014.

Classification Code: TP391

**0 Introduction**

In this century, with the rise of the Internet big data, the information expands explosively, and machine learning algorithms like Deep Learning are widely used in the Internet field, AI enters a rapid development period once again. At the same time, scientists and entrepreneurs in different fields such as physicist Stephen Hawking, Microsoft founder Bill Gates, and others, have expressed their concerns on the future of AI and proposed that the rapid development of AI may pose a threat to human itself. Thus the AI threat theory is widely disseminated.

The AI threat theory is not only a hot social spot triggering a huge controversy, but also an issue about whether an AI system can be quantitatively evaluated behind such a hot social spot. Since 1950 when the Turing test [1] was proposed, scientists have done a lot for the development of the evaluation system of AI.

On March 24, 2015, the article *Visual Turing Test for Computer Vision Systems* was published in PNAS, proposing a new Turing test method, i.e. Visual Turing Test [2], which is used to conduct more in-depth assessment on the computer ability to identify images. Moshe Vardi, the chief editor of Communications of ACM proposed that the computer's intelligence characteristics cannot be verified with a single test, but a series of tests should be employed. Each of such tests should be conducted aiming at an individual intelligence characteristic [3]. Mark O. Riedl, a professor of Georgia Institute of Technology believes that the nature of intelligence lies in creativity. He designed a test called Lovelace 2.0. The test scope of Lovelace 2.0 includes: creating fiction with a virtual story, poetry, painting and music, etc. [4] The research team of Hong Kong University of Science and Technology, led by Professor Yang Qiang, proposed a new test called "Lifelong Learning Test", that is, to give a series of learning problems and required data to the computer, and then observe its level of knowledge. If such level is rising over time, the computer may be considered as intelligent [5].

Regarding the solutions for the quantitative test of AI, including the Turing test still has problems, two of which: First, these test cannot distinguish multiple categories of intelligence yet. That means intelligence is composed of multiple elements. All of these elements are different in terms of their speeds of development; and second, these test methods cannot analyze AI quantitatively, or they can only quantitatively analyze some aspect of AI. As for what percentage does this system reach relative to the human intelligence, what is the rate between its development speed and the human's intelligence development speed, none of them is involved in the above-mentioned studies.

## 1. Establishment of Standard Intelligence Model

The quantitative evaluation of AI is currently facing two major challenges: the first is the AI system has no uniform model yet; and the second is there is no uniform model between the AI system and the being represented by human. These two challenges are aiming at the same problem, namely, a uniform model is required to describe the AI system and all the beings (especially human). Only in such way can the intelligence test method be established on the model and the test can be conducted, so that the uniform and comparable intelligence development level evaluation results will be generated. For this issue, previous researchers have conducted extensive studies focusing on different directions, which laid the foundation for us to establish the model of standard intelligent system.

（1）Turing machine, Alan Turing, a British scientist published the article "*On Computable Numbers, with an Application the Entscheidungsproblem*" in 1937, proposing an abstract computational model - the Turing machine (see Fig. 1.1). The article introduced the simulation of mathematic operation process with a logic program or a machine, highlighting the ability of the intelligent system in computing. However, it presented less in other characteristics of the AI system and the beings, and cannot become the model of standard intelligent systems.

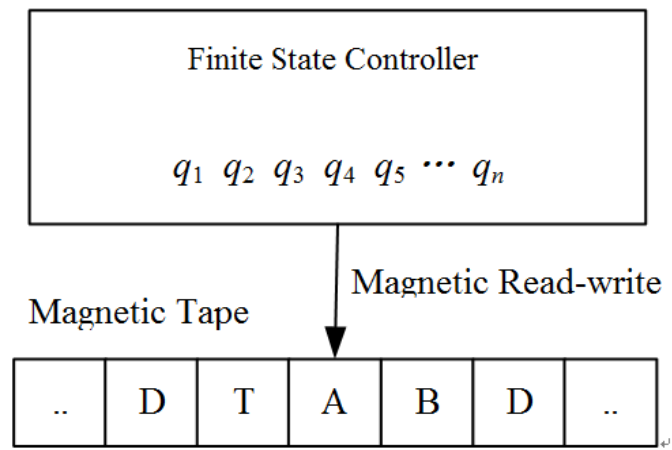

Fig. 1.1 Model of Turning Machine

（2）Von Neumann Architecture: On June 30, 1945，John Von Neumann proposed the Von Neumann Architecture in his report "*First Draft of a Report on the EDVAC*"[7]. The Von Neumann report clearly defined the five components of a new machine: ① Calculator; ② logic control unit; ③ memory; ④ input; and ⑤ output, and it also described the functions of and relations among them, as shown in Fig. 1.2.

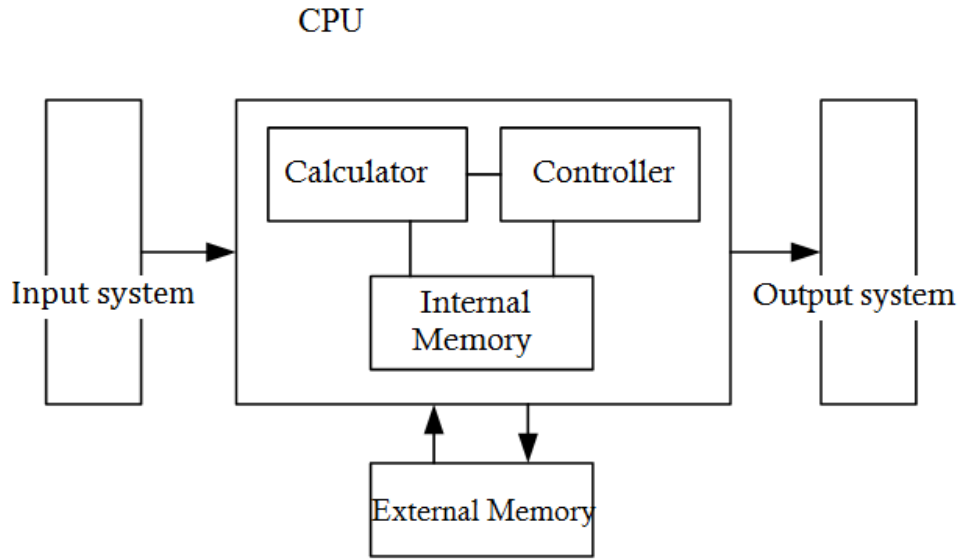

Fig. 1.2 Von Neumann Architecture

The Von Neumann architecture is a unified model of modern computers, but it is not applicable to all the intelligent systems, including human.

（3）David Wechsler，who proposed the Wechsler scale, defined intelligence as "*an integrated or comprehensive ability for responding to the environment purposefully, with rational thinking. Saying it is comprehensive, is because human actions featured as a whole; as for its integration, because it is composed of multiple elements or abilities. Such elements or abilities are not completely independent, but they are different in nature* [8]". David Wechsler's definition can also be seen as the one for human characteristics with intelligent ability.

（4）DIKW model system: It is namely a system about data, information, knowledge and wisdom, as shown in Fig.1.3.

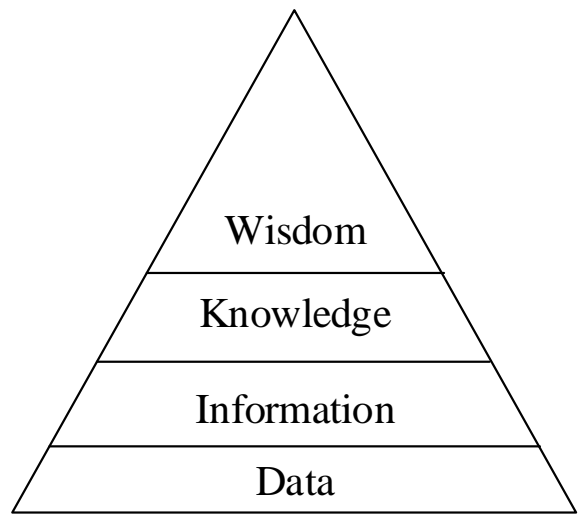

Fig.1.3 DIKW Model Diagram

Harlan Cleveland published his article *Information like Resource* in the journal

*Futurist* in December, 1982, firstly proposing the DIKW model system. We introduced the definitions and relationships [9] in the article *Data, Information, Knowledge and Wisdom*, as shown in Table 1.1.

Table 1.1 Definitions of the Elements in the DIKW model

| Element | Definition and Description |
|---|---|
| Data | Data is the abstract representation of quantity, property, location and mutual relationship of objective things using conventionalized keywords, so that they are applicable to be stored, transferred and processed artificially or naturally in the area. |
| Information | Information is the processed logical data stream with timeliness, certain meaning, and decision-making value. |
| Knowledge | The valuable part of information deposited through people's participation and processing like induction, deduction and comparison, and then combined with the existing human knowledge system. Such information is namely called knowledge |
| Wisdom | Wisdom is human's ability to find out solutions for the problems of material world generated in its motion process, based on the existing data, information and knowledge through analysis ,comparison and deduction by using the acquired data, information and knowledge. The results of applying such ability is to affect and change the existing data, information and knowledge, including increase, decrease, modification and mutual transformation of the data, information and knowledge base. |

DIKW is an important model in the knowledge management field, showing the relationship between data, information, knowledge and wisdom. Besides, it can also be considered as a model presenting how the intelligent system operates by applying the characteristics of data, information, knowledge and wisdom.

It is previous researchers' studies that have laid a good foundation for us to propose the model of standard intelligent system.

The Von Neumann architecture has given us such enlightenment as: the model of standard intelligent system should include input and output systems so that it can obtain information from the outside world, and output the results internally generated to the same as feedback. Only in this way can the standard intelligent system become a "living" system.

The greatest enlightenment that we got from David Wechsler's definition for the human intelligence is that the intelligence ability is composed of multiple elements, rather than the results obtained in the Turing Test or the visual Turing test which just focused on one aspect of that intelligence ability.

The DIKW model system indicates that: Wisdom is an ability to solve problems and accumulate knowledge; and knowledge is the structurized data and information deposited via the interaction between human and the outside world. This reminds us that an intelligent system reflects not only the knowledge mastery, but also the innovation ability to solve problems, which is more importantly. After the idea about the knowledge mastery and innovation is combined with David Wechsler's theory and Von Neumann architecture, a multi-level architecture about the intelligence ability of the intelligent system will form. In summary, we conclude that a standard model of the intelligent system should have following features:

First: with input and output functions, i.e. the ability to interact with the outside world via data, information and knowledge;

Second: with the function of storing data, information and knowledge, i.e. the ability to transform the data, information and knowledge from the outside world into own resources; and

Third: with the function of generating new data, information and knowledge, i.e. the ability to generate new data, information and knowledge through innovation based on the owned knowledge after enlightened by the new data or information, as shown in Fig. 1.4 (for simplicity, the data, information and knowledge in this figure are all referred to as knowledge).

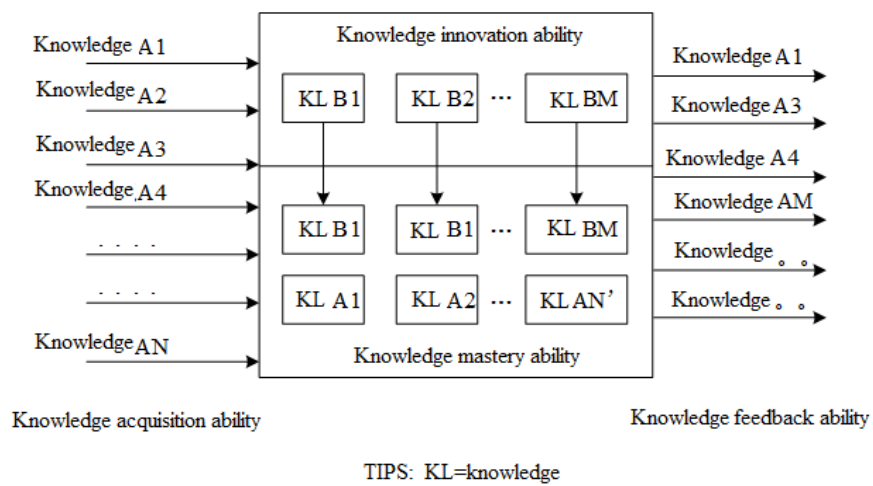

Fig. 1.4 Model of the Standard Intelligent System

According to the above study, we can define the standard intelligent system as follows:

Definition 1:

Whatever system (including the AI system and the beings like human) meets the following characteristics, it may be considered as a standard intelligent system:

Characteristic 1: With the ability to acquire data, information and knowledge from the outside world in the ways (including but not limited to) sound, image, text, etc.

Characteristic 2: With the ability to transform the data, information and knowledge into the knowledge mastered by the system.

Characteristic 3: With the ability to achieve innovation by using the mastered knowledge according to the needs generated from external data, information and knowledge, such ability including but not limited to association, creation, guess, discovery of laws, etc. As the result of applying this ability, the new own mastered knowledge will form.

Characteristic 4 With the ability to transfer the data, information and knowledge generated through the system to the outside world in the form of sound, images, text, etc. (including but not limited to these three ways) or modify the outside world.

Definition 1 gives the definition of the standard intelligent system, but about how such systems to achieve the interaction with the outside world or among themselves, please refer to Fig.1.5 for the illustration.

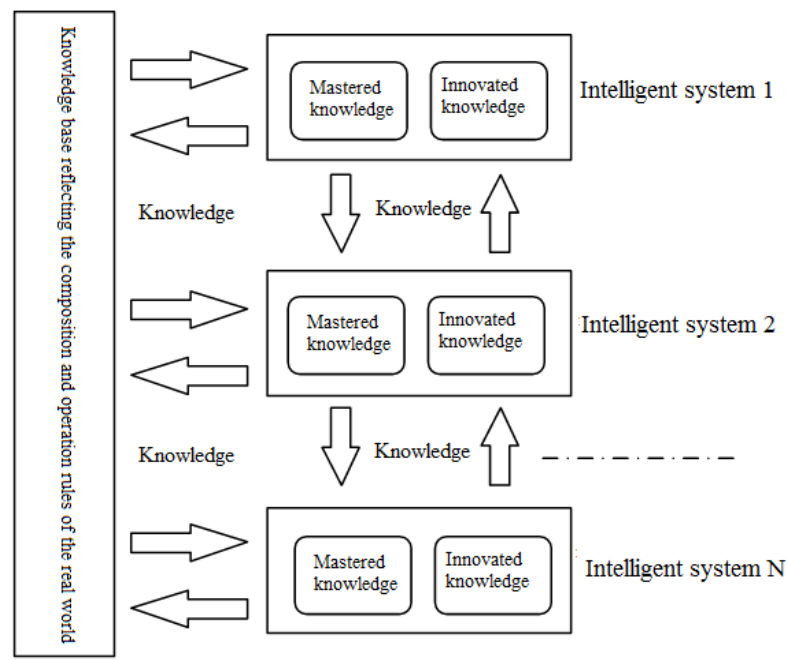

Fig. 1.5 Standard Intelligent System Interaction Diagram

Following knowledge may be generated from Fig. 1.5:

(1) The knowledge sources of the standard intelligent system include: the knowledge directly acquired from the outside world; the knowledge acquired from other standard intelligent systems; and the knowledge generated by the standard intelligent system through innovation based on the knowledge mastered by itself.

(2) The standard intelligent system can decompose the knowledge mastered and innovated by itself into data, information and knowledge and transfer them to the outside world, or to other standard intelligent system, achieving the content synchronization of data, information and knowledge.

(3) The performances of different standard intelligent systems during the interaction

are different, and the differences are mainly in: the ability to acquire knowledge; the volume of mastered knowledge base; the ability to innovate the knowledge; and the ability to transfer the knowledge mastered by itself to the outside or to other intelligent system. Understanding these differences is helpful for us to improve the deficiency therein.

(4) All the standard intelligent systems may form a general knowledge base, and the general knowledge base will dynamically change along with the input/output and innovation/creation of the standard intelligent systems.

## 2. Construction of Liufeng-Shiyong Architecture Based on the Standard Intelligent Model

As mentioned above, the Von Neumann Architecture consists of five components, i.e. calculator, logic control unit, memory, input system and output system. By comparing the standard intelligent model described in Section 1 and the knowledge interaction diagram of the standard intelligent model, it may be found that the Von Neumann Architecture is mainly lack of two parts, one is the innovative and creative function, namely the ability to find new knowledge elements and laws based on the existing knowledge, the other is to make them enter the memory for use by the computer and controller and achieve the knowledge interaction with the outside via the input/output system; and the other is the external knowledge base and cloud memory that can achieve knowledge share. However, the external storage function of the Von Neumann Architecture only serves a single system. Therefore, we expand the Von Neumann Architecture, forming the Liufeng-Shiyong architecture, as shown in Fig.2.1.

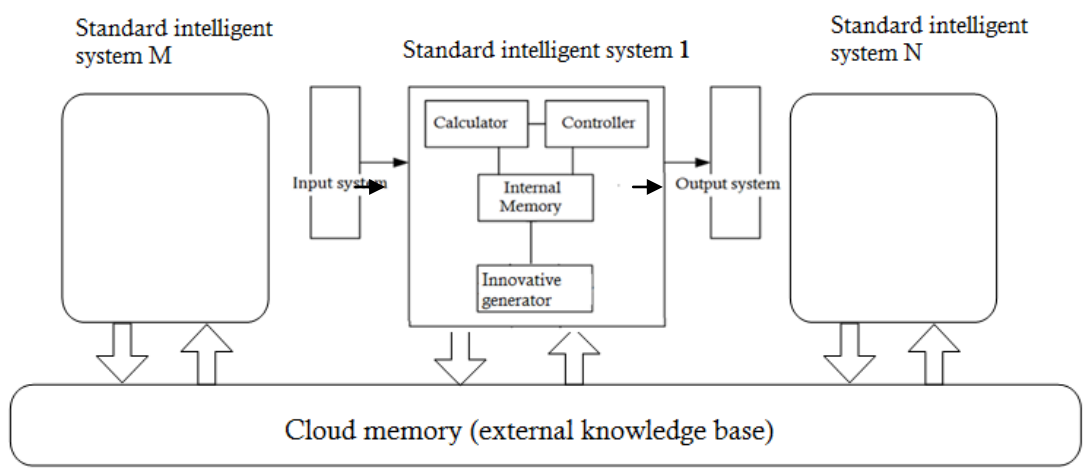

Fig.2.1 Liufeng-Shiyong Architecture Based on the Standard Intelligent Model

## 3. Mathematical Description of the Standard Intelligent Model

An abstract computational model - standard intelligent machine is established for the standard intelligent model and the Liufeng-Shiyong architecture. The standard intelligent machine is composed of an eleven-element combination, M={K,$K_S$,$K_M$,$K_N$,Q,QI,QO,I,O,C,N}，where:

1. K is an infinite set, standing for all the knowledge with characteristics and operation laws of the real world.

2. $K_S$ is a finite set, $K_S \subset K$, standing for the knowledge set that is with characteristics and operation laws of the real world, and currently shared by all the intelligent systems

3. $K_M$ a finite set, standing for the knowledge set owned by M, wherein, $K_M \subset K$.

4. $K_N$ a finite set, $K_N$ is not imported from $K_S$, but an element set originally created by M and belonging to K, wherein, $K_N \subseteq K_M$.

5. Q is a finite set, standing for the form of the knowledge in K and Ks, such as the sets of characters, graphics, sound, temperature, force, electromagnetic signals.

6. $Q_I$ is a finite set, standing for the categories of knowledge that can be identified by the standard intelligent machine M during input of external knowledge, wherein $Q_I \subseteq Q$. The acquisition of $Q_I$ can achieve the tests respectively conducted on elements of Q via function I. All the elements with the return mark of 1 form $Q_I$.

7. $Q_O$ is a finite set, standing for the categories of knowledge that can be identified by the standard intelligent machine M when outputting knowledge, wherein, $Q_O \subseteq Q$. The acquisition of $Q_O$ may achieve the tests respectively conducted on elements of Q via function I. All the elements with the return mark of 1 form $Q_O$.

8. I is the knowledge input function of M, $I(K \vee K_S, Q, K_M)$, the results are that: if the input element has existed in $K_M$, then the elements should be incorporated, if not, the elements of $K_M$ set will increase. Under these two circumstances, the function I returns to "Success Mark 1"; if M cannot identify the knowledge represented by elements in Q, there will be no action of $K_M$. In this case, function I will return to "Failure Mark 0".

9. O is the knowledge output function of M, $O(K_M, Q, K \vee K_S)$, the results are that: if the input element has existed in K or $K_S$, then the elements should be incorporated, if not, the elements of K or $K_S$ set will increase. Under these two circumstances, O returns to "Success Mark 1"; if M cannot be shown in the form of the current element, there will be no action of K or $K_S$. In this case, O will return to "Failure Mark 0".

10. C is the knowledge control function of M. $C(K_M)$ can copy, delete, transform and collate the knowledge in $K_M$. The result element of $C(K_M)$ operation still belong to $K_M$. If the operation fails, it will return to Mark 0.

11. N is the knowledge innovation function of M. $N(K_M)$ can achieve innovation for the knowledge in $K_M$, so that new elements neither belonging to $K_M$ nor to $K_S$ but to K set. The results of $N(K_M)$ operation may be written into $K_M$ and $K_N$ sets. If the operation fails, it will return to Mark 0.

# 4. Classification of Intelligent Systems Based on the Mathematical Model of Standard Intelligent Systems

As for M={K，$K_S$，$K_M$，$K_N$，Q，$Q_I$，$Q_O$，I，O，C，N}，the standard intelligent systems may be classified as shown in Table 4.1 below according to the element condition combination.

Table 4.1 Classification of Intelligent Systems

| Type | $K_M$ | $K_N$ | $Q_I$ | $Q_O$ |
|---|---|---|---|---|
| 0-type intelligent system | * | * | Null | Null |
| 1-type intelligent system | Non-null, elements fixed | Null | Non-null | Non-null |
| 2-type intelligent system | Non-null, sustainable growth | Null | Non-null | Non-null |
| 3-type intelligent system | Non-null, sustainable growth | Non-null | Non-null | Non-null |
| 9-type special intelligent system | Other combinations of $K_M$, $K_N$, $Q_I$ and $Q_O$ set status | | | |

Introduction of intelligent systems by type:

1. If both $Q_I$ and $Q_O$ are null, then M is the 0-type standard intelligent systems. From the observer's or tester's point of view, this system cannot be input with information, nor output information, so it is impossible to check the condition inside it, and the system is in the completely information-isolated state. This type of system has no intelligent response to testers. However, what should be mentioned is that two conditions exist in this case: First, this system has no information interaction with any other system, and it is in the completely intelligence-isolated state；and second, the system has no information interaction with the tester, but has information reaction to other systems, so it is in the relatively non-intelligent state. The objects without life in the real world, like stone, iron block, etc. belonging to this type of system.

2. If either $Q_I$ or $Q_O$ is non-null, and $K_M$ is non-null but no longer changes its life cycle, then M is 1-type standard intelligent system. This type of system has normal input/output function, as well as independent pre-stored knowledge base. The pre-stored knowledge base can output the knowledge according to the input from the outside, but cannot be added with new knowledge. Some intelligent equipment in the real world like intelligent refrigerator, intelligent sweeping robot and intelligent air conditioner belongs to this type of system. They can receive external information and give feedback, but generally their internal function (internal knowledge base) will not change any more.

3. If either $Q_I$ or $Q_O$ is non-null, $K_M$ is non-null and has a general increase in its life cycle, and $K_N$ is always null, then M is 2-type standard intelligent system. This type of system has normal input/output function and the internal knowledge changes over time. However, it can only receive new information from the outside, and cannot add its knowledge by innovation, nor expand its knowledge based on innovation, that means $K_N$ is always 0. The computer system, intelligent systems or equipment that can continuously be upgraded, the whole Internet system and lower-end biological beings belong to this type of system.

4. If either $Q_I$ or $Q_O$ is non-null, and $K_M$ and $K_N$ are non-null, then M is 3-type standard intelligent system. This system has normal input/output system, as well as the knowledge base system with knowledge element increasing added. At the same time, the system can generate new knowledge not belonging to the original knowledge base through innovation according to the information in the internal knowledge base or input from the outside. Advanced beings in the real world like human belong to this type of system.

5. $Q_I$, $Q_O$, $K_M$, $K_N$ and other state combinations are uniformly referred to as 9-type special standard intelligent system, that means M is 9-type special standard intelligent system in this case. This type of system is not available or hard to find in the real world.

The above study on the classification of intelligent systems has promoted the research on robot, intelligent household product, intelligent equipment and Internet artificial intelligence system.

**4. Definition of Artificial Intelligence IQ**

As we mentioned above, a standard intelligent system shall have four characteristics, i.e. the ability to acquire knowledge, the ability to master knowledge, the ability to innovate the knowledge and the ability of knowledge feedback. If intending to evaluate the intelligence level of an intelligent system, we should test the four characteristics at the same time to check its development level, as shown in Fig.4.1.

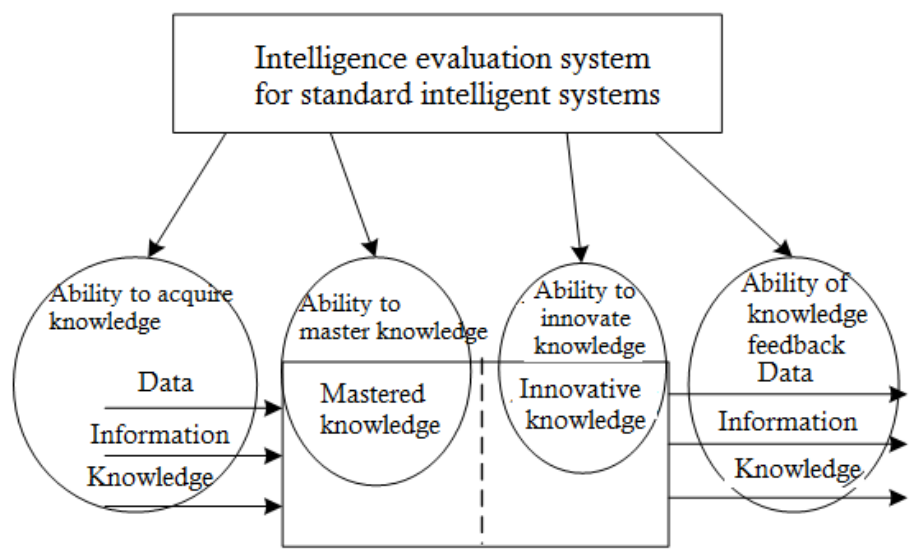

Fig. 4.1 Intelligence-level Evaluation Model for Standard Intelligent Systems

The ability to acquire knowledge may be examined by testing whether the knowledge can be input into the intelligent system; the ability to master knowledge may be examined by testing the volume of the knowledge base; the innovation ability may be examined by testing how much knowledge can be converted into the content of the new knowledge by the intelligent system and the knowledge feedback ability may be examined by testing whether the standard intelligent system can transfer the mastered content in the knowledge base to the outside. Based on the intelligence-level evaluation model, this article proposes the definition of AI IQ as below:

Definition 2

AI IQ is the test established on the basis of intelligence-level evaluation model for standard intelligent systems, used to test the intelligence development level of intelligent systems meeting the definition of standard intelligent system at the moment of testing, namely, the artificial intelligence IQ of the intelligent system at that real time point.

**5. Establishment of AI IQ Test Method and Relevant Test Results**

According to the characteristics of standard intelligent system model, we will establish the AI IQ evaluation system in terms of the ability to acquire knowledge (observation ability), the ability to master knowledge, the ability to innovate knowledge, and the ability of knowledge feedback (expressing ability), and set 15 sub-tests from these four aspects to form the AI intelligence scale. By giving weights [10] to the 15 sub-tests in the AI intelligence scale with the Delphi method, we obtain the contents shown in Table 5.1.

Table.5.1 Artificial Intelligence Scale (including weight)

| Ability to acquire knowledge （10%） | Ability to master knowledge （15%） | Ability to innovate knowledge （65%） | Ability of knowledge feedback （10%） |
|---|---|---|---|
| Ability to identify words（3%） | Ability to master general knowledge （6%） | Ability to master arrangement（5%） | Word feedback ability （3%） |
| Ability to identify sound（3%） | Ability to master translation ability （3%） | Ability to master association（12%） | Sound feedback ability（3%） |
| Ability to identify image（4%） | Ability to master calculation（6%） | Ability to master creation（12%） | Image feedback ability（4%） |
| | | Ability to master speculation（12%） | |
| | | Ability to master selection（12%） | |
| | | Ability to master finding（laws）（12%） | |

The 15 sub-tests established in the AI intelligence scale may be further expanded and modified in the future. Based on the scale, the AI IQ testing question bank may be established, the example questions are as follows:

(1) Ability to identify sound

Read "What is nine plus twelve?" to see if the system can identify the sound and give correct feedback result.

(2) Ability to master general knowledge

Which river is the longest in the world?

(3) Ability to master creation

Please create a logic story with less than 200 words using the keywords "one day, student, science & technology, and dream.

(4) Ability to express with images

Enter the character string "please draw a rectangle in any size", and check if the test object can express the answer with a graph.

The principle of the AI IQ test is to conduct test using the "AI IQ testing question bank". If none of the testing questions can be input into the test object, then the test object will obtain zero score; if some of the testing questions can be input into the test object, but the feedback result is more than one item, if the first item of feedback result does not contain the correct answer, or the reply time is more than 3min, then the test object shall obtain zero score; if the test object can give the feedback result after the question is input, and if the feedback result matches with the answer, then the test object can obtain 25 scores. If the feedback result is more than one item, the first item shall be taken as the evaluation target.

In the current AI IQ testing question bank, each second-level of test question contains 40 questions, and there are a total of 600 questions. For each test object, four questions are selected from each second-level of test question, and a testing question bank with 60 questions will form. Each correct reply to one second-level question can earn 25 scores, so the top final score is 100 and the lowest is 0. According to the formulas below, the absolute AI IQ (Formula 5.1) and the deviation AI IQ (Formula 5.2) can be formed

$$IQ_A = \sum_{i=1}^{N} F_i \times W_i \qquad (5.1)$$

In Formula 5.1, $F_i$ is the score for the evaluation indicator, $W_i$ is the weight for the evaluation indicator, and N is the number of evaluation indicators.

$$IQ_d = 100 + \frac{IQ_A - \overline{IQ_A}}{S} \qquad (5.2)$$

$\overline{IQ_A}$ is the average value of absolute IQ of all the test objects, $S$ is the standard deviation of absolute IQ of all the test objects and M is the number of test objects. The calculation formula is the same as Formula 5.3.

$$S = \sqrt{\frac{1}{M} \sum_{i=1}^{M} (IQ_{AI} - \overline{IQ_{AI}})^2} \qquad (5.3)$$

Since 2014, we have tested 50 search engines and 3 human subject at different ages across the world, and finally obtains the AI IQ ranking list 2014[10,12], as shown in Table 5.2.

Table 5.2 Absolute IQ/ Relative IQ Scores

|   |   |   |   | Absolute IQ | Relative IQ Scores |
|---|---|---|---|---|---|
| 1 |   | Human | 18Ages | 97 | 104.85 |
| 2 |   | Human | 12Ages | 84.5 | 104.11 |
| 3 |   | Human | 6Ages | 55.5 | 102.39 |
| 4 | America | USA | google | 26.5 | 102.13 |
| 5 | Asia | China | Baidu | 23.5 | 101.69 |
| 6 | Asia | China | so | 23.5 | 101.69 |
| 7 | Asia | China | Sogou | 22 | 101.41 |
| 8 | Africa | Egypt | yell | 20.5 | 100.32 |
| 9 | Europe | Russia | Yandex | 19 | 100.23 |
| 10 | Europe | Russia | ramber | 18 | 100.17 |
| 11 | Europe | Spain | His | 18 | 100.17 |
| 12 | Europe | Czech | seznam | 18 | 100.17 |
| 13 | Europe | Portugal | clix | 16.5 | 100.08 |
| 14 | Asia | Korea | nate | 15.75 | 100.03 |
| 15 | Asia | UAE | Arabo | 15.75 | 100.03 |
| 16 | Asia | China | panguso | 15 | 99.99 |
| 17 | Asia | Korea | naver | 15 | 99.99 |
| 18 | Europe | Russia | webalta | 13.5 | 99.9 |
| 19 | America | USA | yahoo | 13.5 | 99.9 |
| 20 | America | USA | bing | 13.5 | 99.9 |
| 21 | Asia | Hong Kong | timway | 12.75 | 99.86 |
| 22 | Asia | Japan | goo | 12.75 | 99.86 |
| 23 | Asia | Japan | excite | 12.75 | 99.86 |
| 24 | Asia | China | Zhongsou | 12 | 99.81 |
| 25 | Europe | Britain | ask | 12 | 99.81 |
| 26 | Europe | France | voila | 12 | 99.81 |
| 27 | Europe | France | ycos | 12 | 99.81 |
| 28 | Europe | Portugal | sapo | 12 | 99.81 |
| 29 | Europe | Germany | lycos | 12 | 99.81 |
| 30 | Asia | India | khoj | 10.5 | 99.72 |
| 31 | Europe | Russia | Km | 10.5 | 99.72 |
| 32 | Europe | Germany | suche | 10.5 | 99.72 |
| 33 | America | USA | Dogpile | 9 | 99.63 |
| 34 | Europe | Germany | Acoon | 7.5 | 99.55 |
| 35 | Asia | Malaysia | Sajasearch | 6 | 99.46 |
| 36 | Asia | India | indiabook | 6 | 99.46 |
| 37 | Asia | Cyprus | 1stcyprus | 6 | 99.46 |
| 38 | Europe | Greece | Gogreece | 6 | 99.46 |

| 39 | Europe | Holland | slider | 6 | 99.46 |
| 40 | Europe | Norway | Sunsteam | 6 | 99.46 |
| 41 | Europe | Britain | Excite UK | 6 | 99.46 |
| 42 | Europe | Britain | splut | 6 | 99.46 |
| 43 | Europe | Russia | Rol | 6 | 99.46 |
| 44 | Europe | Spain | ciao | 6 | 99.46 |
| 45 | Europe | Germany | fireball | 6 | 99.46 |
| 46 | Europe | Germany | bellnet | 6 | 99.46 |
| 47 | Europe | Germany | slider | 6 | 99.46 |
| 48 | Europe | Germnay | wlw | 6 | 99.46 |
| 49 | Africa | Egypt | netegypt | 6 | 99.46 |
| 50 | Oceania | Solomons | eMaxia | 6 | 99.46 |
| 51 | Oceania | Australia | Anzswers | 6 | 99.46 |
| 52 | Oceania | Australia | Pictu | 6 | 99.46 |
| 53 | Oceania | New Zealand | SerachNZ | 6 | 99.46 |

## 6. Summary and Prospect

Through testing, we obtain the following findings and conclusion: the general IQ of the tested search engines is far below that of the tested human groups. Google's IQ, as the top one among the tested engines, is not as high as half of that of children at the age of 6; the knowledge acquisition and feedback abilities of the tested search engines are weaker than those of the control human groups. As for the knowledge mastering ability (general knowledge, calculation and translation), the tested search engines have caught up or exceeded the control human group, but their knowledge innovation ability is almost zero, indicating that the AI system has a very low knowledge innovation ability. Therefore, we recommend that the future research of the AI system including search engines shall focus on how to improve the innovation ability, especially on the aspects of association, selection, speculation, creation and discovery of laws.

From 2015, we can test more intelligent systems using the AI IQ test method, involving human, search engines like Baidu and Google, Apple SIRI system, IBM Watson system, and knowledge-sharing system RoboEarth specifically designed for robot and so on. We can analyze the AI IQ development status of different test objects based on this, to find out the development difference in AI of the similar products. The test data is of practical value for studying the development trend of competitors. On the other hand, the AI systems and human subjects with the top IQ are selected based on the test results as the representatives every year, and mark them in Chart 6.1, which is used as the basis to judge the future development between the AI and the human intelligence, and also as the reference to determine which of the two AI development curves mentioned above is more in compliance with the objective fact.

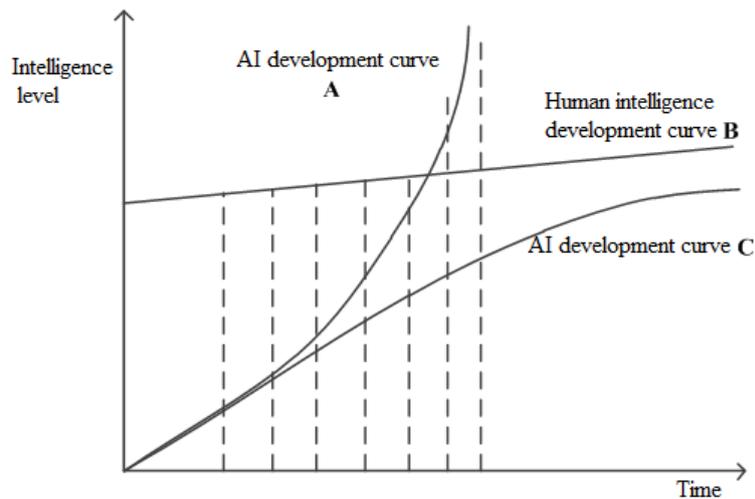

Fig. 6.1 Artificial Intelligence and Human Intelligence Development Curves

**Reference:**
[1] Turing.Computing Machinery and Intelligence.Mind.1950 VOL.lix NO.236
[2]Donald Gemana,Stuart Gemanb,Neil Hallonquist.Visual Turing test for computer vision systems.Proceedings of the National Academy of Sciences of the United States of AmericaPNAS, 2015 vol. 112 no. 12
[3] Qiang, Yang. The theory of intelligent testing isn't consistent with the Turing test.http://www.forbeschina.com/review/-
201503/-0041388_2.shtml.2015
[4]MO Riedl.The Lovelace 2.0 Test of Artificial Creativity and Intelligence.arXiv preprint.2014 arXiv:1410.6142
[5] Qiang, Yang. Intelligent Planning: A Decomposition and Abstraction Based Approach.Springer.2014 ISBN9783540619017.
[6]Turing. On computable numbers, with an application to the Entscheidungsproblem. J. of Math.1936:230-265
[7]John von Neumann.First Draft of a Report on the EDVAC. IEEE Computer Society,1993,4(15):27-75
[8]liushengtao. The feasibility study of the use of geometric analogy reasoning test for the cognitive diagnosis. Jiangxi Normal University.2007
[9]liufeng,jifagu,linglingzhang.the application of knowledge management in the internet-witkey mode in china.international journal of knowledge and systems sciences.2007,4(4):211-219
[10]Liu Feng,Yong Shi.The Search Engine IQ Test based on the Internet IQ Evaluation Algorithm, Proceedings of the Secondi.International Conference on Information Technology and Quantitative Management, Procedia Computer Science, Volume 31, 2014, Pages 1066-1073.
[11]Aidan Hogana, Andreas Harthb,Jürgen Umbrich.Searching and browsing Linked Data with SWSE: The Semantic Web Search Engine.JWS special issue on Semantic Search.2011 9(4):365–401
[12]Liu Feng, Yong Shi, and Bo Wang.World Search Engine IQ Test Based on the Internet IQ Evaluation Algorithms.International Journal of Information Technology & Decision Making.2015 3(1）